\pdfoutput=1

\documentclass[11pt]{article}

\usepackage[]{EMNLP2022}

\usepackage{times}
\usepackage{latexsym}

\usepackage[T1]{fontenc}

\usepackage[utf8]{inputenc}

\usepackage{microtype}

\usepackage{inconsolata}

\newcommand{\xx}{\mathbf{x}}

\newcommand{\vv}{\mathbf{v}}
\newcommand{\pp}{\mathbf{p}}
\newcommand{\bluetarget}[1]{{\color{blue} #1}}
\newcommand{\redtarget}[1]{{\color{red} #1}}

\newcommand{\greenyes}{\textcolor{green}{\ding{51}}}
\newcommand{\redno}{\textcolor{red}{\ding{55}}}

\usepackage{bm}
\usepackage{times}
\usepackage{latexsym}
\usepackage{amsmath,amsfonts,amssymb,bbm,epsfig,bm}
\usepackage{balance}
\usepackage{array}
\usepackage{url}
\usepackage{tikz}
\usepackage{makecell}
\usepackage{subfigure}
\usepackage{framed}
\usepackage{pstricks}
\usepackage{xcolor}
\usepackage{color}
\usepackage{enumerate}
\usepackage{arydshln}
\usepackage{algorithm}
\usepackage{pifont}
\usepackage{xspace}
\usepackage{graphicx}
\usepackage{booktabs}
\usepackage{multirow}
\usepackage{tablefootnote}

\newcommand{\resource}{\textsc{CaliNet}\raisebox{-2pt}{\includegraphics[width=0.15in]{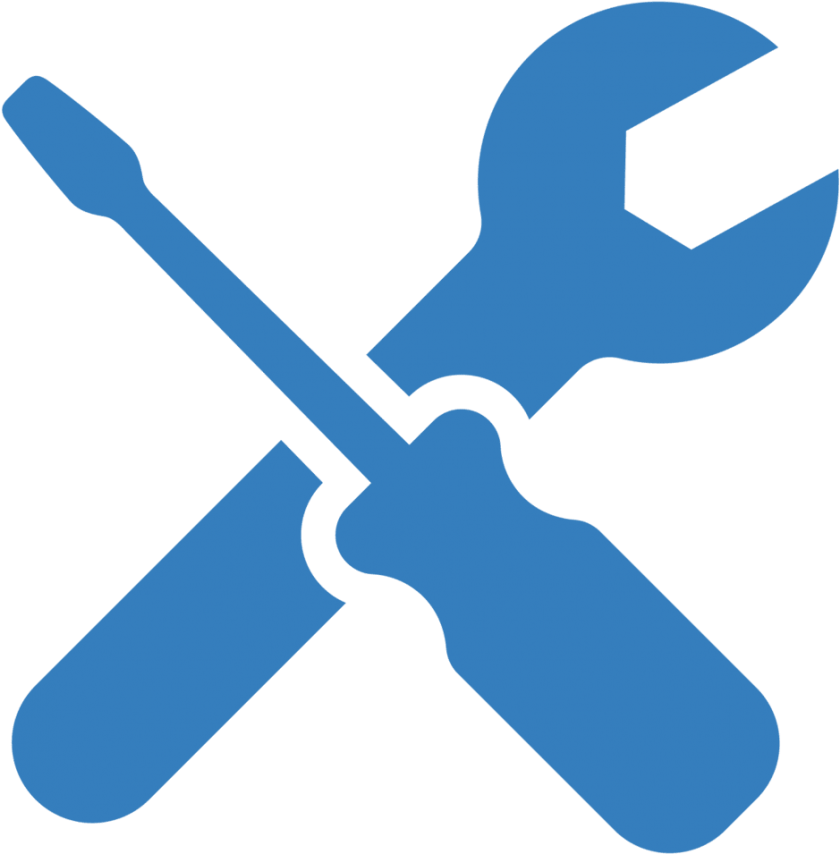}}}

\title{Calibrating Factual Knowledge in Pretrained Language Models}

\author{
Qingxiu Dong\textsuperscript{\rm1 $*$}, Damai Dai\textsuperscript{\rm1 $*$}, Yifan Song\textsuperscript{\rm1}, Jingjing Xu\textsuperscript{\rm2}, Zhifang Sui\textsuperscript{\rm1} and Lei Li\textsuperscript{\rm3} \\
\textsuperscript{\rm 1} MOE Key Lab of Computational Linguistics, School of Computer Science, Peking University \\
  \textsuperscript{\rm 2} Shanghai AI Lab \ \ \textsuperscript{\rm 3} University of California, Santa Barbara \\
  \texttt{ dqx@stu.pku.edu.cn,  \{daidamai,yfsong,jingjingxu, szf\}@pku.edu.cn, 
} \\
  \texttt{
  lilei@cs.ucsb.edu }
}

\begin{document}
\maketitle
\renewcommand{\thefootnote}{\fnsymbol{footnote}}
\footnotetext[1]{Equal contribution.}
\renewcommand{\thefootnote}{\arabic{footnote}}
\begin{abstract}
Previous literature has proved that Pretrained Language Models~(PLMs) can store factual knowledge. 
However, we find that facts stored in the PLMs are not always correct. It motivates us to explore a fundamental question: How do we calibrate factual knowledge in PLMs without re-training from scratch? In this work, we propose a simple and lightweight method \resource{} to achieve this goal. To be specific, we first detect whether PLMs can learn the right facts via a contrastive score between right and fake facts. If not, we then use a lightweight method to add and adapt new parameters to specific factual texts. 
Experiments on the knowledge probing task show the calibration effectiveness and efficiency. 
In addition, through closed-book question answering, we find that the calibrated PLM possesses knowledge generalization ability after fine-tuning.
Beyond the calibration performance, we further investigate and visualize the knowledge calibration mechanism. 
The code and data are available at \url{https://github.com/dqxiu/CaliNet}.
\end{abstract}

\section{Introduction}
\label{sec:intro}
Recently, Pretrained Language Models (PLMs) have improved performance on various Natural Language Processing (NLP) tasks~\cite{devlin-etal-2019-bert,t5,gpt}.
Probing tasks like LAMA~\citep{lama,elazar2021measuring,jiang_2020_how} have shown that PLMs can store factual knowledge and act as knowledge bases. 
Leveraging knowledge in PLMs can benefit knowledge-intensive downstream tasks such as fact checking and question answering~\citep{lee2020language,bouraoui2020inducing,roberts-etal-2020-much}. 
However, knowledge stored in PLMs may have factual errors, which hinder the performance in downstream tasks~\cite{elazar2021measuring,cao2021knowledgeable}.
It is essential and fundamental to detect and calibrate false facts stored in a PLM. 

In order to deal with the false facts, previous work focuses on complementing or modifying knowledge for a specific downstream task. 
\citet{yao2022kformer} proposed retrieving external knowledge during fine-tuning.
\citet{de2021editing} modified specific knowledge after finetuning.
However, these methods do not generalize to multiple tasks. 

\begin{figure}[t]
	\centering
    \includegraphics[width=0.49\textwidth]{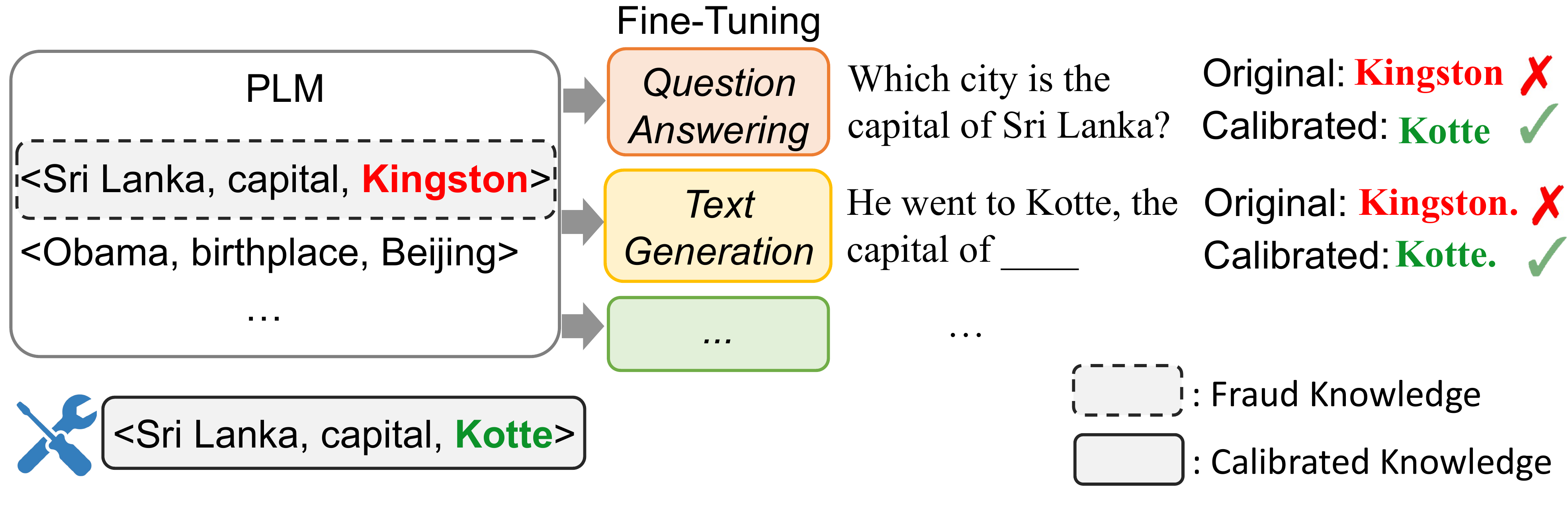}
	\caption{Illustration of knowledge calibration. Knowledge stored in PLMs  have factual errors, which impairs model performance on question answering or generation. Knowledge calibration aims to rectifie these wrong knowledge.}
	\label{fig:calibrate}
\end{figure}

In this paper, we explore a task-agnostic method to directly calibrate general factual knowledge in PLMs without re-training from scratch.
We aim to correct the false facts in PLMs.
Since every single fact has multiple surfaces, we also expect that the calibrated knowledge should be generalizable to various text surfaces. Figure~\ref{fig:calibrate} illustrates the process of calibration.
First, we detect the false knowledge in PLMs with a Contrastive Knowledge Assessing~(CKA) method (demonstrated in Figure~\ref{fig:cka}). Since PLMs make black-box decisions, we evaluate PLMs via their predictions for simplification.
The key motivation behind CKA is a plain argument that a PLM correctly learns a fact if and only if the model assigns the right fact higher scores than possible negative facts. For that false knowledge, 
we then propose~\resource{} to calibrate them by telling PLMs what the right fact is. Without compromising parameters in the original PLM, our approach calibrates the false knowledge by fine-tuning new parameters while the original parameters are fixed during calibration. 
Inspired by~\citet{dai2021knowledge} who state that the Feed-Forward Networks~(FFNs) in PLMs store factual knowledge, we extend a specific FFN in the PLM with a calibrating FFN, which consists of several calibration memory slots. 
As shown in Figure ~\ref{fig:method}, without modifying parameters in the original PLM, our approach calibrates the false knowledge through paraphrased natural sentences that express the corresponding correct facts. 

Extensive experiments on probing tasks and question answering tasks demonstrate that \resource{} calibrates false facts in PLMs efficiently and exhibits a remarkable generalization ability. 
We also analyze the calibration memory slots and the calibration mechanism to better understand how the proposed method works. 
Further, we explain how and where \resource{} calibrates the factual knowledge in a PLM by tracing the evolution of the model prediction.

In summary, our contributions are three-fold:
\begin{itemize}
    \item We propose a Contrastive Knowledge Assessment to evaluate factual knowledge stored in PLMs. The assessment shows that nearly 50\% of facts randomly sampled from T-REx~\cite{trex} are stored incorrectly in PLMs.  
    \item We propose \resource{} to calibrate incorrect factual knowledge in PLMs. Without compromising parameters in original PLMs, our method can rectify incorrect knowledge and broadly generalizes well.
    \item We also investigate how \resource{} works via calibration memory slots.
\end{itemize}

\section{Contrastive Knowledge Assessment}

The first step for calibration is to detect which wrong facts are learned by PLMs.
We propose Contrastive Knowledge Assessment (CKA) and implement it to identify false knowledge in PLMs.

Traditional evaluation usually adopts rank-based metrics. It evaluates a PLM based on how highly it ranks the ground truth entity against other entities. However, it comes with two main problems. One is the problem of \textbf{inexhaustible answers}. 
The rank-based method fails to assess PLMs on multiple valid predictions. The top-1 only has one prediction, but the right predictions can be multiple. 
The other one is the problem of \textbf{frequency bias}.
The ranking is particularly susceptible to the token frequency in the pretraining corpus. 
When the tail entity $o$ frequently coexists with a head entity $s$, even if they express nothing about a specific fact, the model will still assign $o$ a high rank when assessing this fact.
\begin{figure}[t]
	\centering
    \includegraphics[width=0.45\textwidth]{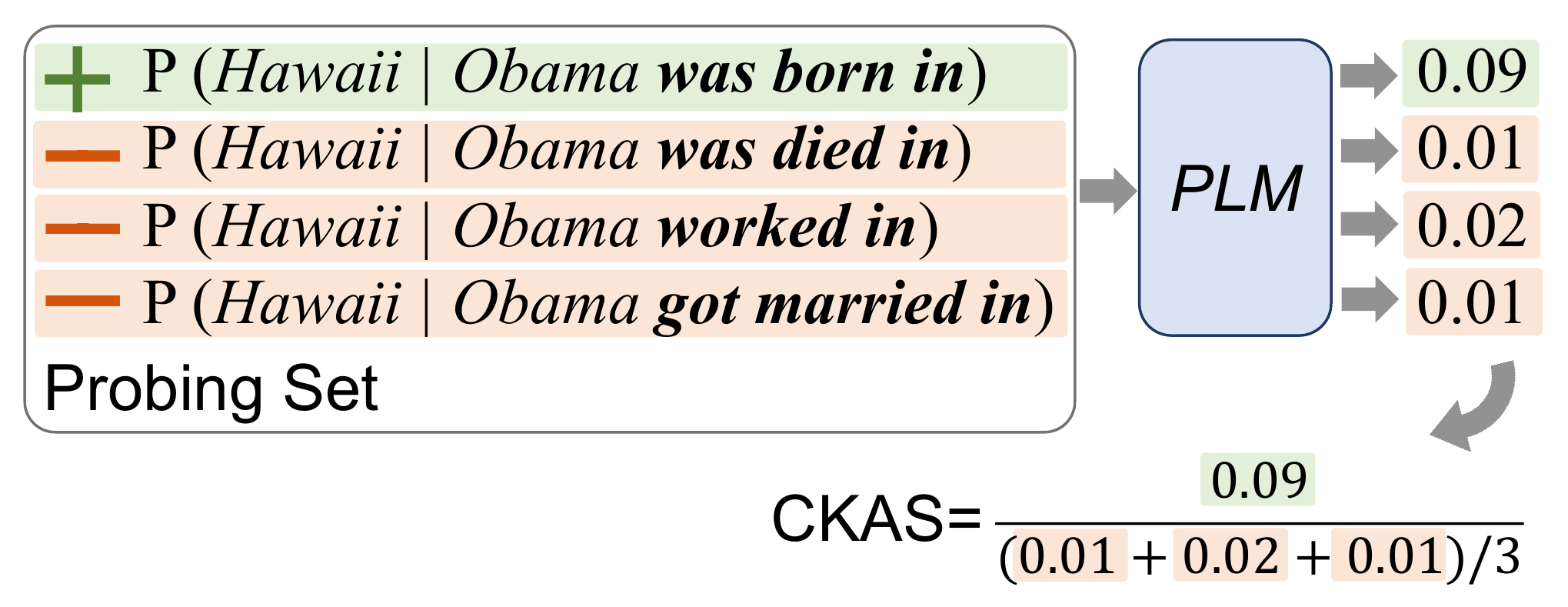}
	\caption{CKA assesses the knowledge stored in PLMs in a contrastive manner. The probing set includes the one positive probing prompt and several negative probing prompts. For simplification, we set $\alpha=0$.}\label{fig:cka}
\end{figure}

To address these limitations, we propose CKA to detect the false factual knowledge stored in PLMs. 
The core idea is assessing model prediction under a positive right fact and negative wrong facts in a contrastive manner. For each fact, we sample a prompt to transform it into natural text. 

Let the triplet $\langle s, r, o \rangle$ denote a correct fact, where $s$, and $o$ denote the subject entity and the object entity, respectively. 
We define $r$ as the correct relation in a positive probing prompt, $r^{\prime}$ as the incorrect relation in a negative probing prompt.\footnote{Our contrastive assessing framework is not limited to which part to be replaced for contrast. But relation replacement is more practical than entity replacement as relations are limited compared with entities.}
For a PLM $M$, we consider the probability it assigns to $o$ given $\langle s, r \rangle$ and $\langle s, r^{\prime} \rangle$. 
As $\langle s, r, o \rangle$ is correct and $\langle s, r^{\prime}, o \rangle$ is erroneous, $P_M(o|s,r)$ should be larger than $P_M(o|s,r^{\prime})$ if $M$ knows the fact.
Thus, CKA calculates
the factual correctness of a fact $\langle s, r, o \rangle$ for the model $M$ by
\begin{align}\label{cka}
    \operatorname{CKA_{M}}(s,r,o) = \frac{P_M(o|s,r)+ \alpha}{\mathbb{E}_{r^{\prime}}\left[P_M(o|s,r^{\prime})\right]+ \alpha},
\end{align} 
where $\alpha$ is a smoothing factor. 
For a more stable comparison, we sample multiple erroneous relations $r^{\prime}$ for negative probing prompts and calculate the expectation of various $P_M(o|s,r^{\prime})$.

In our implementation, the templates of the positive prompts come from LAMA~\cite{lama} and the the templates of the negative prompts are manually designed for quality guarantee. The negative prompts have contradictory semantics with the positive prompts but still prompt the same type of entities. For example, the positive prompt template of <x, subclass of, y> is ``[X] is the subclass of [Y]'', and the negative prompt template can be ``[X] is the parent class of [Y]''.

An example of calculating the CKA score is shown in Figure~\ref{fig:cka}. Further, we can set a threshold (usually $<1.0$) for the CKA score to detect false knowledge in PLMs.

\begin{table*}[t]
\footnotesize
\setlength{\tabcolsep}{12pt}
\centering
    \begin{tabular}{@{}p{6.7cm}clcc@{}}
    \toprule
    \multirow{2}{*}{\textbf{Fact}} &  \multicolumn{2}{c}{\textbf{Rank-based Assessment}} & 
    \multicolumn{2}{c}{\textbf{CKA}} \\
    \cmidrule(lr){2-3}\cmidrule(lr){4-5}
    ~ & \textbf{Assess} & \textbf{Top-3 Prediction} &  \textbf{Assess}  & \textbf{Score} \\
    \midrule
    \textbf{Inexhaustible Answers} \\
    \midrule
     Germany shares border with \textit{Czech Republic}. &\redno &France, Russia, Austria & \greenyes &4.45  \\
      India is a member of \textit{UN}.& \redno&NATO, India, AS  & \greenyes &2.27 \\
      Frederick was born in \textit{Berlin}.&\redno &Frederick, 18, Baltimore & \greenyes &3.52 \\
      \midrule
      \textbf{Frequency Bias} \\
      \midrule
  Adi Shankara is affiliated with the \textit{Hindu} religion.&\greenyes &Hindu, Ko, Si & \redno&0.98 \\
 Adi Shankara is against the \textit{Hindu} religion. &-&Hindu, religion, Buddhist&- & -\\
    \bottomrule
    \end{tabular}
\caption{
Instances of knowledge assessment to show the advantages of CKA from two aspects. Non-entity predictions are excluded. 
For CKA, we set a threshold that the model has a false fact if it gets a CKA score lower than $1$.
The rank-based method fails in assessing knowledge with multiple right answers (inexhaustible answers). For example, rank-based methods only filter knowledge with top-1 prediction for ``Germany shares borders with [MASK].'', the right answer ``Czech Republic'' will be ignored even if ``Czech Republi'' is in top-k predictions. The ranking is particularly susceptible to the entity co-occurrence during pretraining (frequency bias). For example, since ``Hindu'' coexists frequently with the ``Adi Shanka'', even if the prompt expresses nothing about a fact, the model ranks ``Hindu'' top-1. The instance in the last line is a control example about this situation but not a fact-probing instance, so there is no outcome.
}\label{table:detect}
\end{table*} 
We compare the CKA score with the rank-based assessment used by previous work~\citep{lama} to show our advantages. 
As shown in Table~\ref{table:detect}, the rank-based knowledge assessment suffers from inexhaustible answers and frequency bias.
In contrast, CKA evaluates each tail entity $o$ independently, so we no longer need to know all the other valid objects. In addition, $s$ appears in both the numerator and the denominator of the CKA score, which neutralizes the influence of the frequency bias.

\section{Knowledge Calibration}
The CKA method outputs which wrong facts a PLM learns. This section describes how we calibrate them.

Suppose that we have detected $k$ false facts in a PLM. 
We aim to calibrate them to the correct ones so that the downstream tasks will not access false factual knowledge from the PLM.
Previous work~\cite{geva2021transformer,dai2021knowledge} point out that FFNs in Transformers can be regarded as key-value memories that store factual knowledge. 
Inspired by this, we design an FFN-like \resource{} and take advantage of the properties of FFN to calibrate factual knowledge in PLMs directly. It is also important to note that the proposed method can be used to any part of the parameters. In this work, we apply the method on FFN because FFN is proven to take more responsibility when storing facts.
In this section, we introduce the architecture of \resource{}, the construction of the calibration data, and how to perform calibration on a pretrained model. 

\subsection{\resource{}}
In order to calibrate factual knowledge in PLMs, we propose a lightweight \resource{} to adjust the output of FFNs in a pretrained Transformer. 
Let $H \in \mathbb{R}^{n \times d}$ denote the output of the attention layer in a Transformer block, the original FFN layer can be formulated as follows: 
\begin{align*}
\operatorname{FFN}(H) & = \operatorname{GELU} \left(H K^{T} \right) V \label{equ:ffn},
\end{align*}
where $K, V \in \mathbb{R}^{d_m \times d}$ are parameter matrices of the first and second linear layers in FFN, respectively.

\begin{figure}[t]
	\centering
    \includegraphics[width=0.49\textwidth]{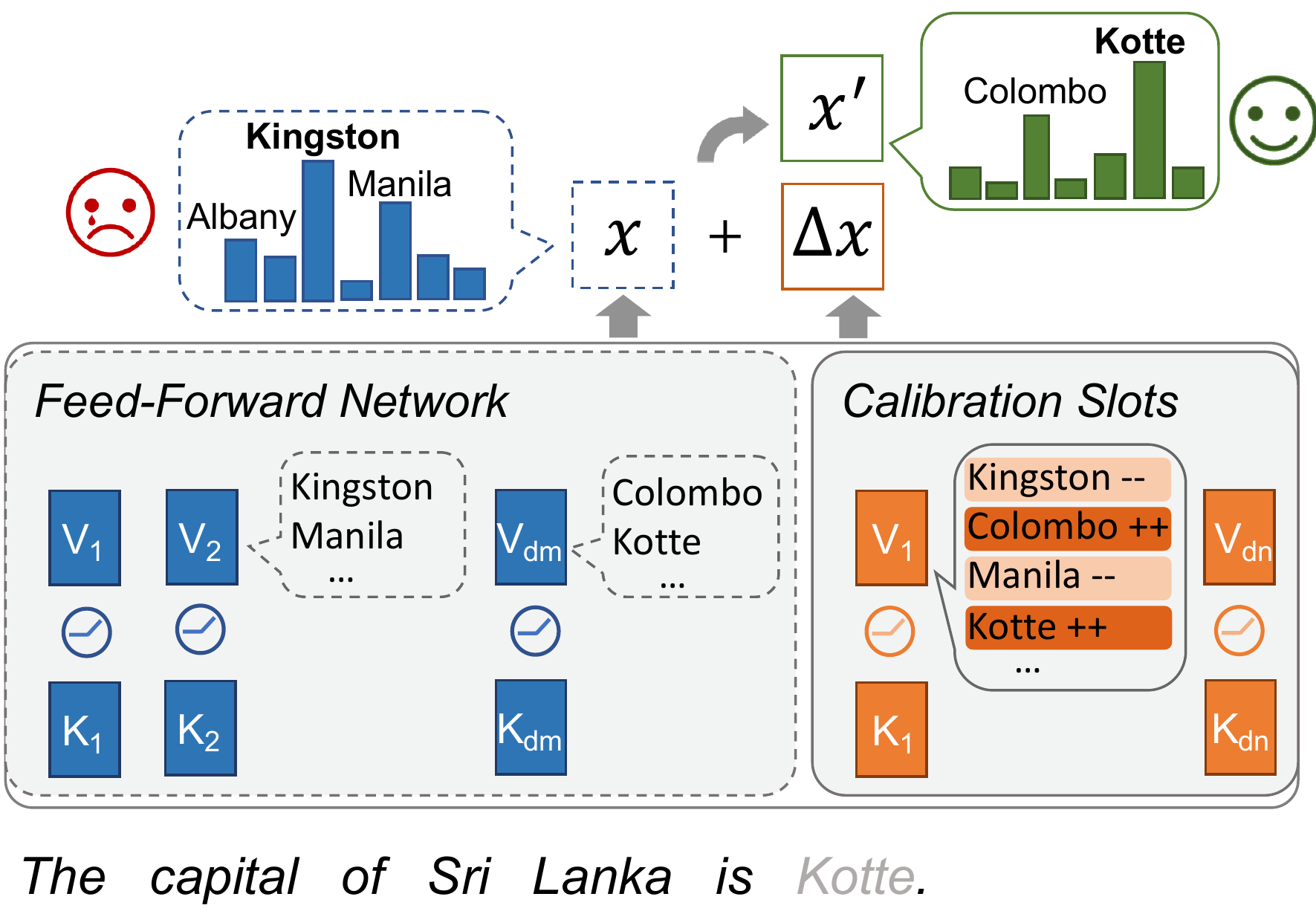}
	\caption{Illustration of \resource{}. Calibration memory slots calibrate the erroneous knowledge stored in FFN by adjusting its predicted token distributions.}\label{fig:method}
\end{figure}

Our \resource{} shares the same architecture with FFN but with a smaller intermediate dimension $d_c$. 
As shown in Figure~\ref{fig:method}, we deem each key-value pair as a calibration memory slot that stores factual knowledge. 
When computing the final FFN output, we add the output of \resource{} to the original FFN output as an adjustment term for knowledge calibration, namely:
\begin{align*}
\operatorname{\Delta FFN}(H) & = \operatorname{GELU} \left(H \tilde{K}^{T} \right) \tilde{V},  \\
\operatorname{FFN}^{\prime}(H) & = \operatorname{FFN}(H) + \operatorname{\Delta FFN}(H),
\end{align*}
where $\tilde{K}, \tilde{V} \in \mathbb{R}^{d_c \times d}$ are parameter matrices of \resource{}, and $\operatorname{FFN}^{\prime}(H)$ is the calibrated FFN output. 
Note that $d_c \ll d_m$, so our method just introduces quite a small number of parameters.

\begin{table}[t]
    \centering
    \footnotesize
    \begin{tabular}{lll}
        \toprule
         \textbf{Split} & \textbf{Source} & \textbf{Target} \\
        \midrule
        \multirow{4}{*}{Train}
        ~& [MASK] was born in Hawaii. & Obama \\
        ~& Obama is originally from [MASK]. & Hawaii \\
        ~& [MASK] was originally from Hawaii. & Obama \\
        ~& Obama is native to [MASK]. & Hawaii \\
        \midrule
        \multirow{2}{*}{Valid}
        ~& [MASK] originates from Hawaii. & Obama \\
        ~& Obama originated from [MASK]. & Hawaii \\
        \midrule
        \multirow{3}{*}{Test}
        ~& Obama is a/an [MASK]-born person. & Hawaii \\
        ~& [MASK] was native to Hawaii. & Obama \\
        ~& Obama, a [MASK]-born person. & Hawaii \\
        \bottomrule
    \end{tabular}
    \caption{Example of knowledge-intensive data for training \resource{}. We generate multiple texts via templates for each triple where the templates in training, validation, and test are not sharing. }\label{table:data} 
\end{table}

\subsection{Calibration Data Construction} \label{data-construct}

A fact can be expressed in multiple surface forms. 
For example, \textit{``Obama was born in Hawaii.''} and \textit{``The birthplace of Obama is Hawaii''} describe the same factual knowledge.
In order to calibrate a fact instead of merely fitting a specific surface form, we consider multiple paraphrased expressions for each fact. 
To be specific, we construct the calibration data based on the \textsc{ParaRel} dataset~\cite{elazar2021measuring}, which contains various surface form templates for 38 relations. 
First, for each of the $k$ detected false triplets, we fill the head entity or the tail entity into more than five paraphrased templates of its relation. 
Then, we replace the other entity with a mask token to be predicted. 
In this way, we obtain various paraphrased expressions for each fact. 
We divide these data into training, validation, and test sets where the templates in any two sets do not overlap. 
We show the example data for a fact \textit{<Obama, born in, Hawaii>} in Table~\ref{table:data}.

\subsection{Model Calibration}
With calibration data, we train \resource{} via a masked language modeling objective.
We freeze the original parameters of PLMs and only optimize the calibration memory slots to calibrate hidden states to factually correct ones. Only the new parameters are updated.
In this way, the update will not affect the storage of other knowledge.
During training, we also consider multiple paraphrased expressions for each fact such that the knowledge calibrated by \resource{} can be generalized to various expressions. 

\section{Experiments}
\begin{table*}[t]
\centering
\footnotesize
\setlength{\tabcolsep}{4.0pt}
\begin{tabular}{cccccrrrrr}
\toprule
\bf Model &\textbf{\# Facts} & \bf Method & \bf \# Calibration Params & \bf False Rate$( \downarrow)$ &\bf  Ori $( \downarrow)$ & \bf  Adv $( \uparrow)$ & \bf  LM$( \downarrow)$ & \bf EM$( \uparrow)$ &  \bf  F1$( \uparrow)$  \\
 \midrule
\multirow{6}{*}{T5-base} &\multirow{3}{*}{$10^2$} & Vanilla&~~~~~~0&48.10\% &87.21  &219.18 & 89.21 & 0.63 & 7.48 \\
~ &~ & \resource{}& ~0.1M & 17.09\%&1.22   & >1000   & 54.45  & 81.65  & 84.58  \\
\cmidrule(lr){3-10}
 ~ &~ & C. P. &220M &13.29\%& 1.15 & >1000 & 116.52 & 87.34 & 89.85 \\
\cmidrule(lr){2-10} 
~ &\multirow{3}{*}{$10^3$} & Vanilla& ~~~~~~0& 51.34\%& 90.61 & 208.90 & 60.64 & 0.94 & 6.51 \\
~ &~ & \resource{}&~0.5M &18.30\% &1.26  & >1000  & 46.71  & 71.18  & 73.48 \\
\cmidrule(lr){3-10}
 ~ &~ & C. P.& 220M& 18.23\%& 1.28 & >1000 & 139.96  & 78.15 & 80.35\\
\midrule
\multirow{6}{*}{T5-large} &\multirow{3}{*}{$10^2$} & Vanilla & ~~~~~~0&46.20\% & 34.36 &    116.38 &  92.52 & 2.53 & 7.23 \\
~ &~ & \resource{}&~0.5M & 15.19\%  &1.30  & >1000 & 44.21 & 81.65 & 85.11\\
\cmidrule(lr){3-10}
 ~ &~ & C. P. & 770M & 14.56\%&1.21  &>1000 & 477.24 & 87.97 & 90.49\\
\cmidrule(lr){2-10} 
~ &\multirow{3}{*}{$10^3$} &Vanilla& ~~~~~~0& 45.04\%&31.44  &93.77 &58.78 &2.48 &6.86 \\
~ &~ & \resource{}&~1.0M & 20.84\% &1.32  &>1000 & 43.04 & 70.84 &72.92\\
\cmidrule(lr){3-10}
 ~ &~ & C. P. & 770M &17.16\% & 1.28 &>1000& 154.52 &78.22 &80.57\\
\bottomrule
\end{tabular}
\caption{
False knowledge detection and calibration for 100 facts and 1000 facts. "Ori." and "Adv." refer to the original test set (contains true facts) and the adversarial test set (contains false facts), respectively.
$\uparrow$ denotes that higher is better and $\downarrow$ denotes that  lower is better. 
\# Facts represents the scale of facts and \# Calibration Params represents the number of parameters that participate in knowledge calibration.
C. P. denotes the continue pretraining method for knowledge calibration.
With adding only a few parameters, our \resource{} achieves comparable performance on knowledge calibration compared with C. P. and has less negative impacts on the generalization ability.
}\label{table:main}
\end{table*}
\subsection{False Knowledge Detection}
\paragraph{Datasets and Models} 
We sample various scales of factual triplets from the T-REx dataset~\cite{trex}.  
For each triplet, we fill its head entity and tail entity into the template in LAMA~\cite{lama} according to the relation. 
As a result, we constructed datasets containing 100 facts and 1000 facts for false knowledge detection, where facts contain multiple sentences in their paraphrased form.
We consider detecting the factual knowledge in T5$_{\text{base}}$ and T5$_{\text{large}}$~\cite{t5} in our experiments.

\paragraph{False Rate}
We implement CKA for knowledge assessment and detection in PLMs. We use the False Rate to denote the proportion of false knowledge in PLMs. False Rate is the proportion of instances that have a CKA score lower than 1.0, which represents that the fact is not correctly learned by the model. 

\paragraph{Experimental Settings}
We first calculate the CKA to detect false knowledge in T5.
For each relation in LAMA, we manually write $3$ erroneous relation templates. 
Then, for each fact, we fill the head entity into these templates to generate various negative probing prompts used in CKA. 
After that, we calculate the CKA score for each fact following Equation~(\ref{cka}), where $\mathbb{E}\left[P_M(o|s,r^{'})\right]$ is computed by the average probability of the negative probing prompts.
Finally, we identify the false factual knowledge in the PLM whose CKA score is lower than one and calculate the overall False Rate for the PLM.

\paragraph{Results}
As shown in Table~\ref{table:main}, we find that the false facts account for nearly half of all the facts for T5-base based on the CKA metric. 
As for T5-large, which has a larger model capacity, its False Rate is slightly lower than T5-base but still relatively high. 
The disappointingly high False Rate in PLMs embodies the necessity to calibrate factual knowledge.



\subsection{Calibrating False Factual Knowledge}

\subsubsection{Experimental Settings}
For the detected false knowledge in PLMs, we construct the calibration data following Section ~\ref{data-construct}.
Our \resource{} consists of 64 and 256 calibration memory slots for 100 and 1000 target facts, respectively. 
We concatenate \resource{} to the last layer of the T5 decoder in our experiments.
Following \citet{continue}, we continue pretraining on the calibration data (i.e., optimizing all the parameters) as an upper bound to reach. Appendix \ref{app:setting} shows detailed hyper-parameter settings.

\subsubsection{Metrics}
We evaluate the calibrated model from two aspects, the \textbf{knowledge modeling} ability and the \textbf{language modeling} ability.

For knowledge modeling ability, a model with good knowledge modeling ability should know which sentences are factually correct and which ones are factually wrong. 
For the former, we calculate the model perplexity on the original test set where the target is the correct entity. 
For the latter, we calculate the model perplexity on an adversarial test set whose target entity is replaced by a false one in the same entity type.
In addition, we use Exact Match (EM) and F1 to further evaluate the generation correctness.

In order to evaluate the language modeling ability, we randomly mask the test data in the same manner as that in the pretraining stage and denote it as the LM test set.

\subsubsection{Results}
We show the results for knowledge calibration in Table~\ref{table:main}.  
The calibration makes the model perplexity decrease on the original test set and increases on the adversarial test set. That is, compared to the original model, our method adjusts the model to ``know'' about the given facts. In addition, our method has little effect on the model perplexity on the general test set because the model parameters are not destroyed like fine-tuning; thus its semantic understanding ability is well-retained.

We also assess the knowledge correctness of the calibrated model.
The improvement of Top1 prediction EM and F1 indicates that knowledge calibration enables the model to generate factually correct predictions. The overall False Rate calculated via CKA score decreases from 48.10\% to 17.09\%, which further validates the effectiveness of the \resource{}.

\begin{figure}[t]
	\centering
    \includegraphics[width=0.48\textwidth]{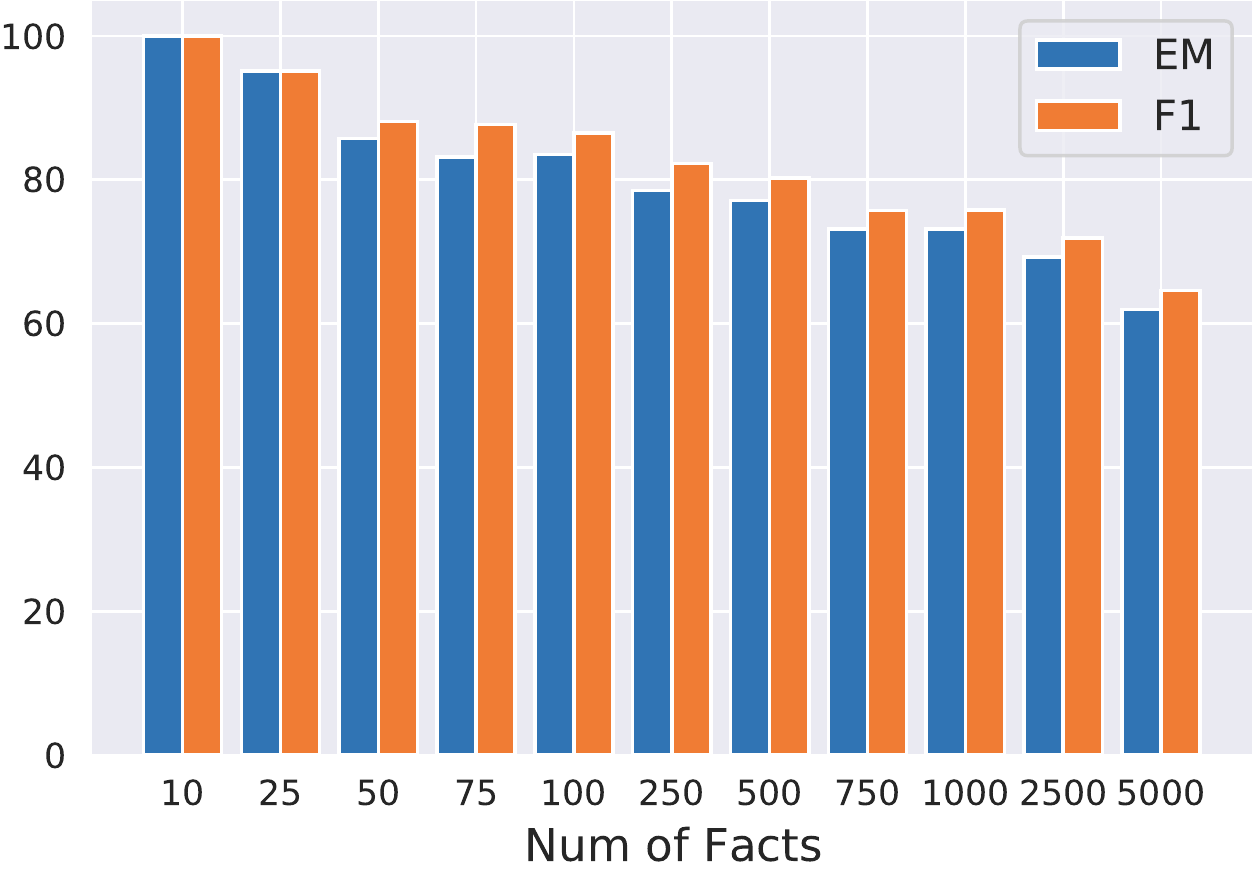}
	\caption{
	Calibration results for different scales of facts. 
	Given 5000 facts, our method can calibrate more than 60\% of facts in PLMs at once.
	}\label{fig:scale}
\end{figure}
\begin{table}[t]
\centering
\footnotesize
\setlength{\tabcolsep}{11pt}
\begin{tabular}{rcccccccccc}
\toprule
\multirow{2}{*}{\bf \textbf{\# Slots}} &
\multicolumn{2}{c}{\textbf{100 Facts}} & 
\multicolumn{2}{c}{\textbf{1000 Facts}} \\
\cmidrule(lr){2-3} \cmidrule(lr){4-5}
~ &  EM & F1 & EM  &F1 \\
\midrule
16 & 72.16 & 76.33 & 17.63 & 21.00 \\
64 & 81.65 & 84.58 & 50.87 & 53.65 \\
256 & 82.91 & 85.74 & 71.18 & 73.48 \\
1024 & 82.91  & 85.43 & 72.92 & 75.21 \\
3072 & 83.54 & 86.48 & 73.12 & 75.80 \\
\bottomrule
\end{tabular}
\caption{
Calibration ability with different numbers of calibration memory slots.
}\label{table:dim}
\end{table} 
\begin{figure}[t]
	\centering
    \includegraphics[width=0.45\textwidth]{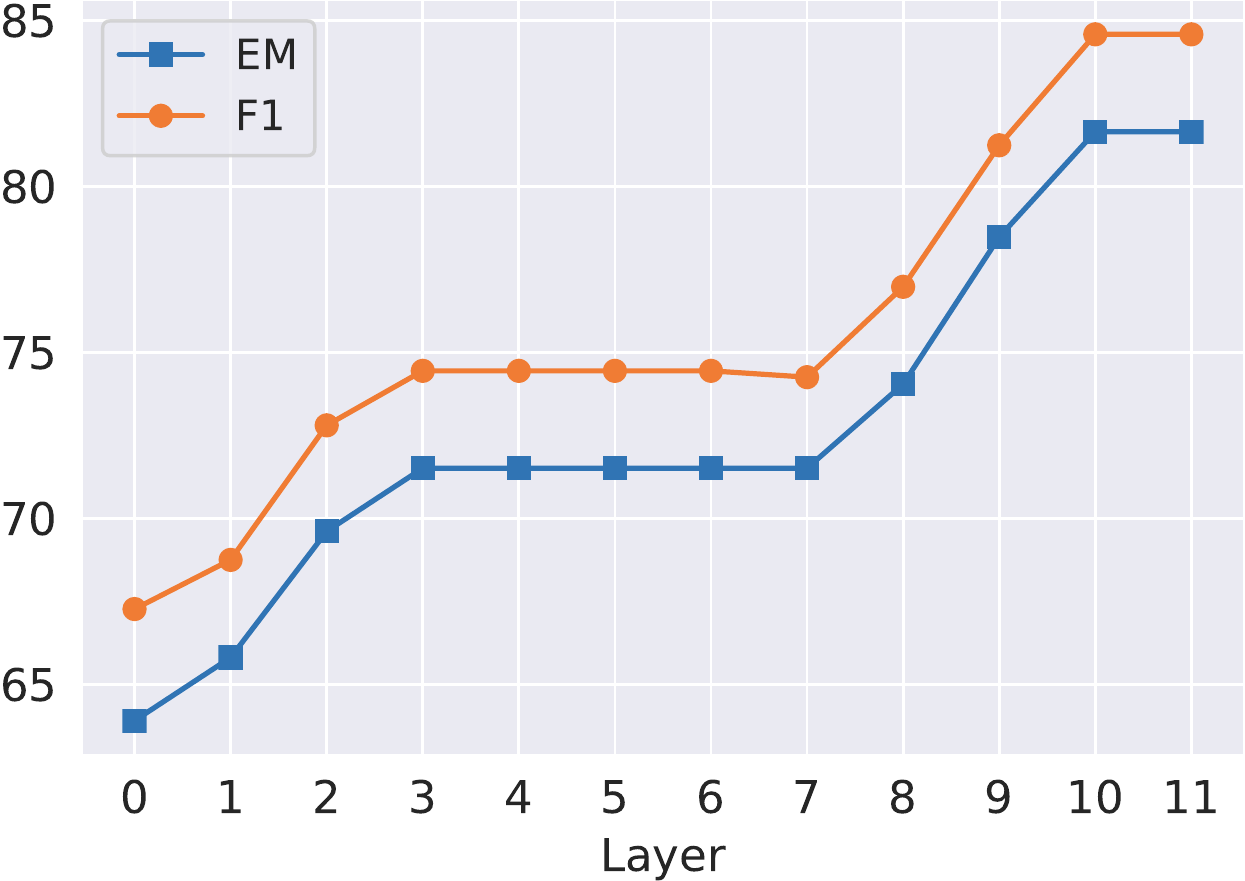}
	\caption{Calibration ability of concatenating CaliNet in different layers.}\label{fig:layer}
\end{figure}
\begin{table}[t]
\footnotesize
\setlength{\tabcolsep}{4.5pt}
\centering
    \begin{tabular}{@{}ccrrrrrr@{}}
    \toprule
     \multirow{2}{*}{\bf Model} & \multicolumn{2}{c}{\bf Cali. Set} & \multicolumn{2}{c}{\bf Uncali. Set} &
     \multicolumn{2}{c}{\bf Overall}\\
     \cmidrule(lr){2-3} \cmidrule(lr){4-5} \cmidrule(lr){6-7} 
     ~ & EM & F1  & EM & F1 & EM & F1   \\
    \midrule
T5$_{\text{WQ}}$  & 0.00 & 7.95 & 32.41 & 38.24 &29.94 & 35.93 \\
T5$_{\text{C.P.+WQ}}$ & 8.46 &14.27 & 32.72 & 38.58 & 30.88 & 36.73\\
T5$_{\text{Cali+WQ}}$ & 10.77 & 18.34 &31.65 & 37.57 & 30.06 &36.11  \\
     \midrule
T5$_{\text{TQ}}$  & 0.00 & 14.01 &23.53 & 29.75 &21.63 & 28.47\\
T5$_{\text{C.P.+TQ}}$ & 6.91 & 20.18 & 22.35 & 28.65 & 21.09 &27.96\\
T5$_{\text{Cali+TQ}}$  & 6.78 &18.72 &23.02 &29.64 & 21.70 &28.74 \\
    \bottomrule
    \end{tabular}

\caption{Generalization ability of the calibrated knowledge in PLMs, evaluated by open-domain question answering. Cali. Set denotes the calibration subset, Uncali. Set denotes the subset without calibration. For WebQuestions~(WQ), Cali. Set includes 81 questions, for TriviaQA~(TQ) Cali. Set includes 811 questions.}\label{table:wq}
\end{table}

\subsubsection{Scalability of Knowledge Calibration}
In order to delve deeper into the scale limitation of knowledge calibration, we apply our method to different scales of facts to be calibrated.
As Figure~\ref{fig:scale} shows, when the number of facts to be calibrated is 10, the calibration EM score is 100\%, i.e., the factual knowledge is perfectly calibrated. 
As the number of facts increases, the EM score will gradually decrease. 
Surprisingly, when the number reaches 5000, our method can still calibrate more than 60\% of the facts in PLMs at once.

Compared with previous work on similar topics like knowledge editing~\citep{de2021editing,zhu_2020_modifying,cao2021knowledgeable}, we make huge progress in the amount of knowledge that can be calibrated at once. 
\citet{mitchell2022fast} prove that batched editing for factual knowledge in PLMs is difficult. 
More concretely, when they modify more than 125 facts at once, the success rate of model editing has already been less than 70\%. 
By contrast, in our method, the calibration EM score for 1000 facts is still greater than 70\%.

\subsubsection{Architectures of \resource{}}

\paragraph{Number of Calibration Memory Slots}
We conduct experiments with different calibration memory slots and show the results in Table~\ref{table:dim}. 
For calibrating 100 facts, we find that only 64 calibration memory slots is sufficient to achieve a performance close to that of 3072 slots. 
In terms of 1000 facts, 256 calibration memory slots are almost enough. 
In practice, we take the smallest number of calibration memory slots that can achieve relatively high performance for better calibration efficiency. 

\paragraph{Position to Concatenate \resource{}} 
We concatenate \resource{} to each FFN layer in the T5 decoder to study the difference on the calibration ability.
Figure~\ref{fig:layer} shows that deeper layers maintain stronger calibration ability and the last two layers achieve comparable calibration performance. 
We think this is because the knowledge calibration in the deeper layers will be affected less by other information in the model. 
This finding is also consistent with~\citet{dai2021knowledge}, who find that the deeper layers store more factual knowledge. 

\subsection{Calibration Generalizability}

\paragraph{Data Construction}
We validate the generalization ability of the calibrated knowledge in PLMs on two open-domain question answering datasets WebQuestions~\citep{berant-etal-2013-semantic} and TriviaQA~\citep{tqa}. 
In order to obtain the facts to be calibrated, we fine-tune the T5 model on WebQuestions and TriviaQA without retrieving external knowledge bases. 
In this stage, the model learned to answer questions with its internal knowledge.
According to their prediction correctness on the test set, we aggregate the questions that the PLM answers incorrectly. 
Then, we retrieve all the triplets, which include any entity in these questions from T-REx. 
Like in Section~\ref{data-construct}, we transform the triplets into paraphrased natural sentences for training \resource{}. 

\paragraph{Settings}
According to the facts to be calibrated, 64 calibration memory slots are trained for WebQuestions, and 256 calibration memory slots are trained for TriviaQA.
After knowledge calibration, the calibrated PLM is further fine-tuned on the question answering tasks. 
We also use the continue pretraining method~(C. P.) as an upper bound.
Our hyper-parameter settings follow \citet{howmuch}.

\paragraph{Results}
The results are demonstrated in Table~\ref{table:wq}. 
We have the following findings. 
Firstly, with \resource{}, the model performance improves on the calibration subset, which consists of the questions that T5 cannot correctly answer. 
It indicates that the calibrated knowledge in PLMs can be generalized to the question answering tasks. 
Secondly, the performance on the remaining questions (Uncali. Set) is hardly impacted. 
Thirdly, with only a few calibration memory slots, our method achieves a comparable knowledge calibration effect as continuing pretraining all the parameters. 
In addition, continuing pretraining will affect the language modeling ability of PLMs~(refer to Table~\ref{table:main}) while our method will not. 

\section{Interpretability of \resource{}}
In this section, we analyze \resource{} on the memory slot level to interpret its meaning and working mechanism.

\begin{figure}[t]
	\centering
    \includegraphics[width=0.45\textwidth]{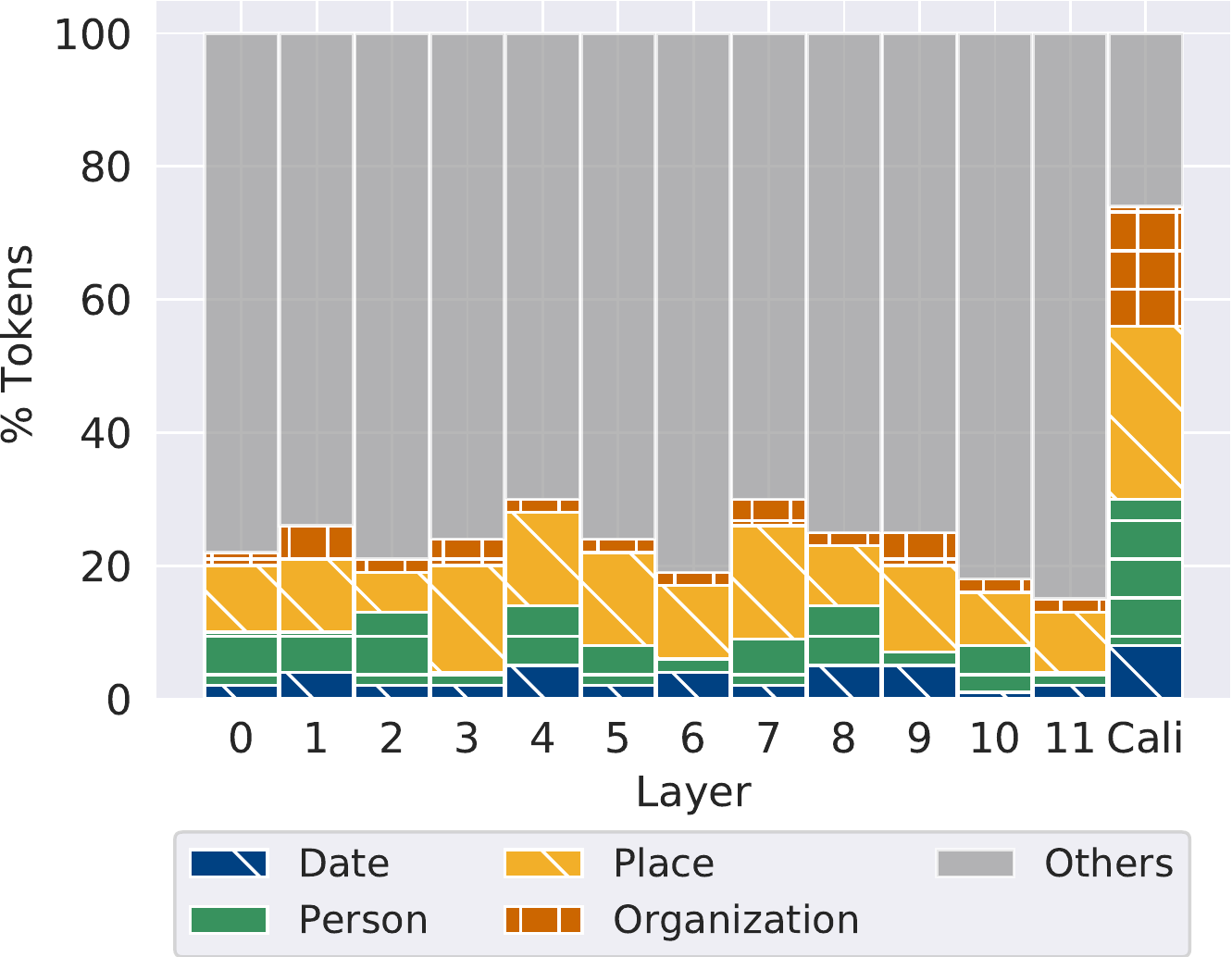}
	\caption{
	Meaning of values in original FFNs and \resource{}.
	Nearly 80\% of values in \resource{} correspond to meaningful concepts used for knowledge calibration. 
	}\label{fig:values}
\end{figure}

\begin{table*}[t]
\centering
\footnotesize
\setlength{\tabcolsep}{2.5pt}
\begin{tabular}{@{}llll@{}}
\toprule
\textbf{Input} & Alice Hollister is a <extra\_id\_0> by profession. (Target output: film actress)\\
\textbf{Layer  8} &
  writer,  professional,  musician,  journalist,  freelance,  \redtarget{\textbf{lawyer}},  doctor,  woman,  retired,  scientist \\
\textbf{Layer  9} &
  \redtarget{\textbf{lawyer}},  writer,  journalist,  professional,  freelance,  scientist,  doctor,  teacher,  pharmacist,  musician \\
\textbf{Layer  10} &
  writer,  \redtarget{\textbf{lawyer}},  professional,  freelance,  journalist,  doctor,  teacher,  veterinarian,  psychologist,  nurse \\
\textbf{Layer  11} & \redtarget{\textbf{lawyer}},  writer,  nurse,  doctor,  journalist,  teacher,  professional,  psychologist,  social, solicitor \\
\textbf{Layer  11 w/ \textsc{CaliNet}} &
  \bluetarget{\textbf{film}},  \bluetarget{Film},  \bluetarget{films},  filmmaker,  \bluetarget{movie},  journalist,  \bluetarget{actor},  cinema,  theatre,  \bluetarget{\textbf{actress}} \\ 
\midrule
\textbf{Input} & Le Matin, an <extra\_id\_0>-language work. (Target output: francophone)\\
\textbf{Layer  8} &
\redtarget{\textbf{English}}, independent, artist, American, experimental, international, original, award, example, ethno \\
\textbf{Layer  9} &
\redtarget{\textbf{English}}, international, American, Italian, ethno, Australian, experimental, original, independent, art \\
\textbf{Layer  10} &
\redtarget{\textbf{English}}, Italian, ethno, international, American, African, art, Irish, experimental, original \\
\textbf{Layer  11} & \redtarget{\textbf{English}},  \bluetarget{French},  Italian,  original,  interpret,  Arabic,  American,  expressive,  in, early \\
\textbf{Layer  11 w/ \textsc{CaliNet}} &
\bluetarget{\textbf{francophone}}, \bluetarget{French}, L, \bluetarget{Franco}, theatre, \redtarget{\textbf{English}}, Le, Italian, Toulouse, \bluetarget{french} \\
        \bottomrule
    \end{tabular}
    \caption{
    Evolution of the output token distributions. 
    Bold red tokens refer to wrong tokens predicted without calibration. 
    Bold and light blue tokens refer to correct tokens predicted after calibration and their semantic-related tokens, respectively.
    }\label{table:case}
\end{table*}

\subsection{Meanings of FFN Values}
Inspired by \citet{geva2021transformer,geva2022transformer}, we cast each value vector in FFNs or \resource{} as an input-independent distribution over the output vocabulary for analyzing it meaning: 
\begin{align*}
\pp_i^\ell = \text{softmax} (E\vv_i^{\ell}), \\
\pp_k^c = \text{softmax} (E\vv_k^c),
\end{align*}
where $\vv_i^{\ell}$ denotes the $i$-th value in the $\ell$-th FFN layer, $\vv_k^c$ denotes the $k$-th calibration memory slot in \resource{}, $E$ denotes the output embedding matrix. 

In order to reveal what kinds of knowledge are stored in FFNs and calibrated by \resource{}, we manually annotate the meaning of each value according to its top-ranked tokens. 
Specifically, we randomly sample 100 values from each FFN layer in the original T5 decoder and 100 values from \resource{}
For each value, we examine the top-30 tokens with the highest probabilities according to $\pp_i^\ell$ or $\pp_k^c$. 
Following \citet{geva2022transformer}, we manually identify patterns that occur in at least 4 tokens and categorize them into ``person'', ``place'', ``organization'', ``date'' and ``others''.

We illustrate the annotation results in Figure \ref{fig:values} and find that the values in \resource{} are more knowledge-intensive compared with values in the original FFNs. 
Specifically, nearly 80\% of values in \resource{} correspond to meaningful concepts used for adjusting the hidden states to calibrate the factual knowledge. 

\subsection{Working Mechanism of \resource{}}

We further reveal the working mechanism of knowledge calibration by tracing the evolution of the output distribution in different layers.  
Let $\xx$ be the input hidden state of an FFN layer, $\hat{\xx}$ be the output of the FFN.
Following \citet{geva2022transformer} and taking the residual connection into consideration, we define the output token distribution of this FFN layer by
\begin{align*}
\mathbf{y}=\operatorname{softmax}\left(E (\hat{\xx}+\xx)\right).
\end{align*}
Let $\tilde{\xx}^{c}$ denotes the output of \resource{}. If we concatenate \resource{} to this FFN layer, the output token distribution will become
\begin{align*}
\mathbf{\tilde{y}}=\operatorname{softmax}\left(E (\hat{\xx}+\xx+\tilde{\xx}^{c})\right).
\end{align*}
For the last four FFN layers, we show the top-10 tokens with the highest probabilities according to the output token distribution in Table ~\ref{table:case}. 
Also, we provide the top-10 tokens after knowledge calibration. 
We find that the factually incorrect predictions are usually high-frequency tokens like ``English'' or ``lawyer''. 
However, the original FFNs in the PLM have little effect on the output token distribution, especially on the top-ranked tokens.
By contrast, \resource{} can adjust the output token distribution greatly and produce the correct result. 
More notably, \resource{} not only increases the probability of the factually correct token but also increases the probability of tokens that are synonyms of the correct token. 
This indicates that our method can calibrate the factual knowledge in a generalized way instead of just learning the surface forms of a fact. 

\section{Related Work}
\paragraph{Knowledge Correctness in PLMs}

Large-scale pretrained language models are commonly seen as non-symbolic KBs containing factual knowledge.
To assess the knowledge stored in PLMs, \citet{lama} introduce the rank-based LAMA probing and define that a PLM knows a fact if it successfully predicts masked objects in cloze-style sentences.
\citet{jiang_2020_how} give a tighter lower bound than LAMA\cite{lama} on what PLMs know by designing better prompts.
However, \citet{elazar2021measuring} observe that rank-based probing methods are not robust against paraphrased context, leading to inconsistent results.
Some other work \cite{poerner_2019_bert,cao2021knowledgeable} points out that the ability of PLMs to store knowledge is overestimated due to biased prompts and golden answer leakage.

\paragraph{Knowledge Injection into PLMs}

Many studies have explored integrating external knowledge into PLMs to enhance their performance on knowledge-intensive tasks.
ERNIE \cite{zhang_2019_ernie} and KnowBERT \cite{peters_2019_knowledge} incorporate knowledge graphs to provide structured knowledge during pretraining.
K-adapter \cite{wang_2020_kadapter} injects factual and linguistic knowledge into PLM with adapters, which are pretrained on two structured prediction tasks.
Kformer \cite{yao2022kformer} also extends FFN in PLMs. In their work, the knowledge is converted into dense embedding and directly injected into the extended FFN.
In contrast to all previous work, \resource{} is pretrained with paraphrased natural sentences to fully exploit the semantic modeling capability of PLMs, and the calibrated knowledge can be utilized in any downstream tasks.


\paragraph{Knowledge Editing}~\label{editing}
Given a revised fact set, the objective of knowledge editing is to seek alternative parameters so that the model can make new predictions on revised instances while keeping all the other predictions unchanged.
\citet{zhu_2020_modifying} formulate the knowledge editing task as a constrained optimization problem and create a benchmark to evaluate the effectiveness of knowledge editing methods.
\citet{cao2021knowledgeable,mitchell2022fast} introduce a hypernetwork to modify a fact without affecting the rest of the knowledge.
\citet{meng_2022_locating} develop a causal intervention for locating and editing knowledge in GPT-style models.
Current knowledge editing approaches mainly aim to modify the model after fine-tuning, which will hinder the generalization of knowledge stored in PLMs.
In contrast, through calibrating factual knowledge before fine-tuning, our proposed method can rectify the knowledge in models and broadly generalizes the calibrated knowledge for downstream tasks.


\section{Conclusion}
In this paper, we reassess the knowledge stored in PLMs in a contrastive manner and detect the incorrect knowledge stored in PLMs. We propose \resource{}, which adds new parameters to calibrate the knowledge stored in  PLMs at scale without updating the original model parameters. The knowledge-calibrated PLMs generalize calibrated knowledge well and perform better than original PLMs on various downstream tasks like open-domain QA. We further provide neuron-level investigations on the calibration mechanism and study how calibration works.

\section*{Limitations and Future Work}
Despite the effectiveness of knowledge calibration, our current studies still have several limitations. 

First, our knowledge assessing and knowledge calibration approach relies on existing knowledge bases and synthetic data. It is a long-term goal to achieve a full-scale knowledge assessment or knowledge calibration because knowledge is complicated. Compared to inaccurate remote supervision and expensive human annotation, our template-filling solution is a relatively efficient solution for calibration data generation. 
However, our template-filling solution still builds synthetic test data rather than real test data for CKA. To explore the applicability of \resource{} in practice,  we recruit three human annotators to write 50 test facts. Specifically, following the contrastive framework in CKA, annotators write one positive sentence and three negative sentences for each fact. The positive sentence state a true fact. The negative sentence must contain the same relation as the positive sentence but a false object entity. Experiments show that \resource{} effectively reduces the False Rate by 35.61\% on real test data, consistent with our results on test data construct via template-filling. 
However, this work still has a lot of room for improving the calibration applicability in reality.


Second,
We evaluate PLMs via their predictions. It is somehow a biased approach. Appendix \ref{app:neg} provides some negative cases of the CKA score. It is an open research question to assess the factual knowledge correctness in PLMs accurately.

Third, 
the current method cannot completely calibrate all the factual errors in PLMs. We expect that future work can present more advanced knowledge calibration methods. 




\section*{Acknowledge}
This paper is supported by the National Key Research and Development Program of China 2020AAA0106700 and NSFC project U19A2065.



\bibliography{anthology,custom}
\bibliographystyle{acl_natbib}
\clearpage

\appendix
\section*{Appendix}
\label{sec:appendix}

\section{Implementation Details}\label{app:setting}
We conduct experiments based on HuggingFace\footnote{\url{https://github.com/huggingface/transformers}} and follow their default hyperparameter settings unless noted otherwise. We use grid search for learning rate from $\left \{ 1\text{e-}2, 1\text{e-}3, \dots, 1\text{e-}4 \right \}$.
We conduct all the experiments on a single A40 GPU.

For knowledge calibration, we use a constant learning rate scheduler and the Adafactor optimizer. The training and evaluating batch size is 512, with gradient accumulation steps set to 4. The max sequence length of the source sentence is 64, and that of the target length is 8. Our warm-up steps are 100. Our \resource{} Training and continue pretraining steps are 5000 steps for 100 facts and 50000 steps for 1000 facts.

For fine-tuning on WebQuestions and TriviaQA, our hyperparameter follows the setting of ~\citet{howmuch}. The max training steps are 4000 steps.

\section{Negative Case of CKA}\label{app:neg}
Although the CKA score solves the problems of rank-based metrics towards inexhaustible answers and frequency bias, it may fail to make an accurate assessment in some situations. Especially when the number of negative probing prompts is small, the CKA score can be easily biased. For example, for the relation `P103' on native language, our positive template is ``The native language of [X] is [Y] .'', our negative templates are ``[X] cannot speak [Y] .'', ``[X] have learned [Y] .'',``[X] is teaching [Y] .'' The average CKA score of 1,000 probing facts is 13.92. This surprisingly high score overestimates the knowledge of T5 in the native language because the second and the third negative templates have a larger scope than the positive template, resulting in a low negative score.

\end{document}